\documentclass{article}
\usepackage{ijcai17}
\usepackage{graphicx,subfigure}
\usepackage{helvet}
\usepackage{amsthm}
\usepackage{amssymb}
\usepackage{amsmath}
\usepackage{multirow}
\usepackage{amsfonts}
\usepackage{graphicx,subfigure}
\usepackage{algorithm}
\usepackage{algorithmic}
\usepackage{courier}
\usepackage{dirtytalk}
\usepackage{lipsum}

\usepackage{times}

\newcommand{\qp}{\ensuremath{Q^{\pi}}}

\title{Reinforcement Learning Based Argument Component Detection}
\author{Yang Gao, Hao Wang, Chen Zhang, Wei Wang\\
Institute of Software, Chinese Academy of Sciences \\
\{gaoyang, wanghao, zhangchen, wangwei2014\}@iscas.ac.cn }

\begin{document}

\maketitle

\begin{abstract}
\emph{Argument component detection} (ACD) 
is an important sub-task in argumentation mining. 
ACD aims at 
detecting and classifying different argument 
components in natural language texts.
\emph{Historical annotations} (HAs) are important 
features the human annotators consider when they
manually perform the ACD task.
However, HAs are largely ignored by 
existing automatic ACD techniques.
\emph{Reinforcement learning} (RL) has proven to be
an effective method for using 
HAs in some natural language processing tasks.
In this work, we propose a RL-based
ACD technique, and evaluate its performance 
on two well-annotated corpora.
%(including a hotel review corpus compiled 
%by ourselves). 
Results suggest that,
in terms of classification accuracy,
HAs-augmented RL outperforms
plain RL by at most 17.85\%, and 
outperforms the state-of-the-art
supervised learning algorithm 
by at most 11.94\%.
\end{abstract}

\section{Introduction}
\label{sec:intro}

The automatic extraction of \emph{arguments} from natural language
texts, also known as \emph{argumentation mining},
has recently become a hot topic in artificial intelligence
(c.f. \cite{lippi2016survey}).
An \emph{argument} is a basic unit people use to 
persuade their audiences to accept
a particular state of affairs \cite{eckle2015role},
and it usually consists of a \emph{claim} and 
some \emph{premises} offered in support of the claim.
As a concrete example,
consider the following texts extracted from a 
hotel review posted on Tripadvisor.com:

\begin{quote}
  \textbf{Example 1}:
  \textcircled{1} Appalling in room television/radio/technology. 
  \textcircled{2} There was an old, small, black, CRT TV. 
  \textcircled{3} The channel selection was minimal,  
  \textcircled{4} picture quality average,
  and \textcircled{5} movie options unimpressive.
\end{quote}

The review excerpt in Example 1 can be viewed as an argument:
clause \textcircled{1} is the claim of the argument, 
and the other four clauses are premises supporting the claim. 
Argumentation mining consists of three sub-tasks {\bf i)} segmenting clauses,
{\bf ii)} distinguishing different argument components 
(e.g. claims, premises) from non-argumentative clauses,
and {\bf iii)} predicting the relations between argument components
(e.g. support/attack).
In this work, we term the second sub-task in argumentation mining
\emph{argument component detection} (ACD), and it is the focus of many
existing argumentation mining papers and this work alike.
%\footnote{
%In this work, we assume that all clauses are provided a priori. 
%There have been some automatic clause segmentation methods,
%e.g. \cite{goudas2014argument,park2015conditional}.}

\noindent \textbf{Motivation.}
When human annotators manually perform the ACD task, 
they decide the label of a clause not only based on
the clause's own linguistic features, but also on its context.
For instance, consider again Example 1:
if we consider clause \textcircled{2} alone and 
ignore its surrounding clauses, we are very likely to 
label it as a claim; 
%as it describes a certain aspect of  the hotel; 
however, if we additionally consider the 
content and label of \textcircled{1}, we may instead label
\textcircled{2} as a premise. 
To obtain the contextual information,
human annotators usually need to read and label a 
document for multiple rounds \cite{ig2014coling}.
However, despite the importance of contextual information,
it is ignored by most existing automatic ACD methods.
%in fact, the confusion between premise and 
%claim is the most significant error in 
%some state-of-the-art AD methods \cite{ig2015emnlp},
%and we believe this is due to the ignorance of
%contextual information.
In this work, we consider a specific form of
contextual information called 
\emph{historical annotations} (HAs), and
investigate how to effectively use
it in ACD.

\begin{figure}
 \includegraphics[width=0.45\textwidth]
 {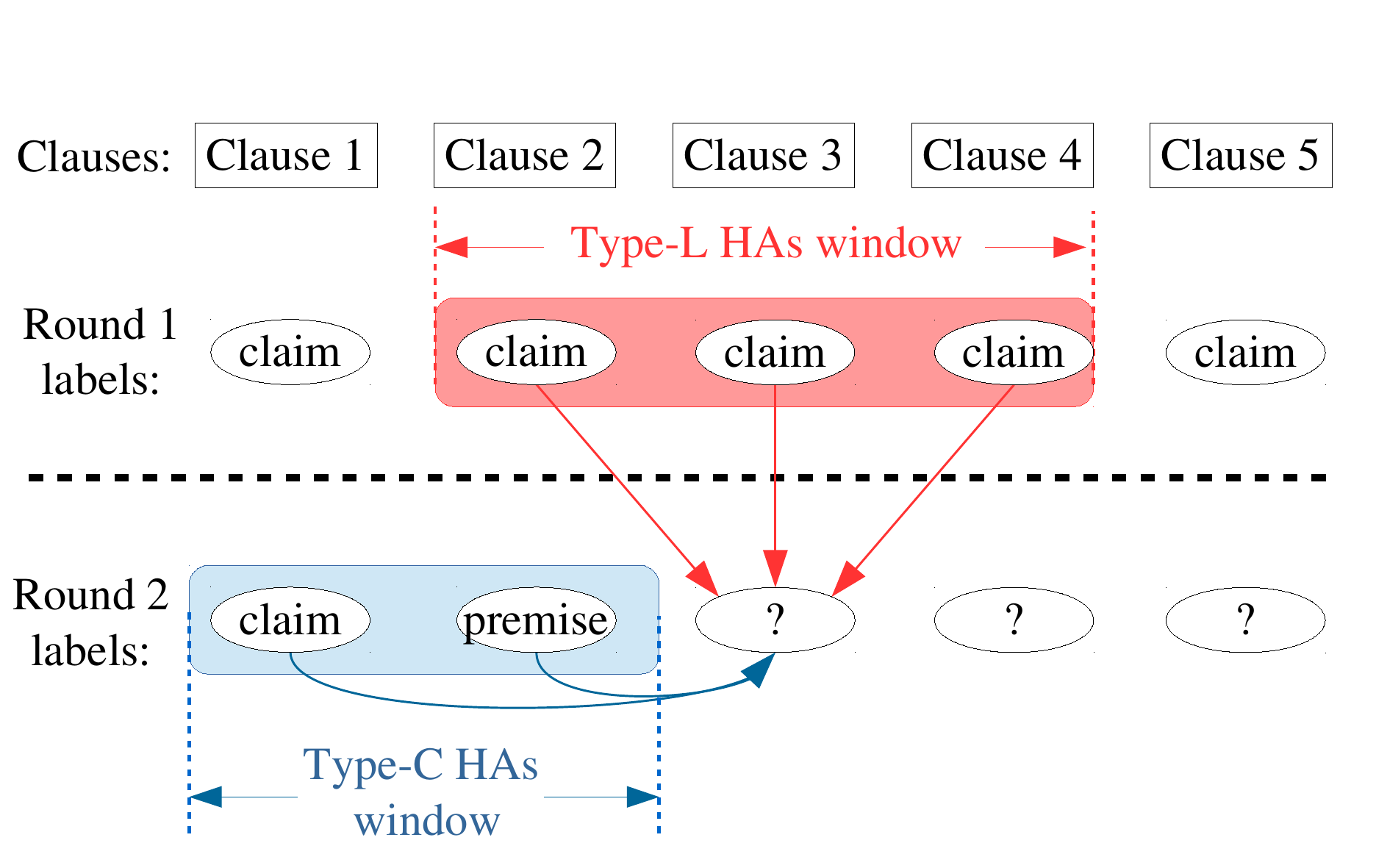}
  \caption{Two types of HAs in ACD.    
  %Type-L and type-C HAs are highlighted in red
  %and blue boxes, resp.
  Note that the labels for the same clause can be different in 
  different rounds of annotating: for example, the label 
  for clause $2$ has been changed from `claim' 
  (in the 1st round) to `premise' (in the 2nd round).
  In this example, the window sizes for type-L
  and type-C HAs are 3 and 2, resp. 
% 
% 
%The \emph{window size} of each type of HAs can be tuned
%so as to control to what extent this type of HAs 
%is considered during the labelling process.
%For example, by letting both types' window size 
%be 0, HAs are ignored in the labelling process.
 \label{fig:has_types}}
\end{figure}

In particular,
given a target clause to be annotated,
human annotators may consider two types of HAs
during their multi-round annotating process:
%since human annotators label a document for multiple rounds,
%%\cite{ig2014coling},
\textbf{type-L} (L stands for 'last round'):
labels of some clauses surrounding the target clause, 
made in the previous round of annotating; and 
\textbf{type-C} (C stands for 'current round'): 
labels of some clauses preceding the target clause,
made in the current round of annotating.
Fig. \ref{fig:has_types} illustrates these two types of HAs.
We consider HAs rather than other types of 
contextual information (e.g. the topic of a
document, linguistic features of some 
surrounding clauses, etc.) for two reasons:
\textbf{i)} 
HAs take only a few bits to encode 
in vectorised representations; and 
\textbf{ii)} 
HAs have been widely used in some NLP
tasks, e.g. text summarisation 
\cite{rioux2014emnlp} and 
dialogue generation \cite{pietquin2011sample}.
To the best of our knowledge, this is the first work
that studies the usage and influence of HAs in the ACD task.
%We leave the study of the other types
%of contextual information as future work.

\noindent \textbf{Objectives.}
%\paragraph{Objectives.}
Our first objective is to present the 
design and implementation 
of the first \emph{reinforcement learning} (RL) based
ACD technique.
When HAs are used in the annotating process, 
the label of the
current clause is part of the contextual information
of surrounding clauses (see Fig. \ref{fig:has_types}); 
thus, the annotating process can be modelled as a 
\emph{sequential decision making} problem,
as the current decision (i.e. label for
the current clause) influences the future decisions.
We formally formulate ACD as a sequential 
decision making problem, and select
suitable RL algorithms to solve it.

Our second objective is to study the influences
of HAs on different ACD methods. 
We evaluate the performances of both RL-based ACD
and some state-of-the-art supervised-learning
(SL) based
ACD on two corpora; results suggest that,
using HAs results in no significant performance
changes for SL-based ACD tools, but 
leads to significant performance improvements
for RL-based ACD; in particular,
by using appropriate HAs, RL's 
accuracy is improved by 8.90\% and 17.85\%
in the two test corpora.
In addition, HAs-augmented RL
significantly outperforms (in terms of accuracy) the state-of-the-art
SL algorithm by 5.56\% and 11.94\% in two test corpora.

\iffalse
Our third objective is to present a well-annotated,
publicly available argumentation
corpora, constructed from 105 hotel reviews
posted on Tripadvisor.com.
Although ACD for product reviews is an interesting
topic for both academia and industrial sectors,
there exist very few argumentation corpora for
product reviews (see Sect. \ref{sec:related_work}).
%ACD in product reviews is not only 
%an interesting research topic in academia,
%but also has great potential for multiple 
%commercial applications \cite{wyner2012semi}.
Our corpus is among the most well-annotated 
argumentation datasets (according to some metrics), 
and we believe it can stimulate and facilitate further 
research on ACD for product reviews.
\fi

%In the rest of the paper,
%we review related work in Sect. \ref{sec:related_work}, 
%give the MDP-based AD formulation in Sect. \ref{sec:mdp_model},
%present the RL-based AD framework in Sect. \ref{sec:select_rl},
%describe the corpora on which we perform our experiments
%in Sect. \ref{sec:data},
%present the empirical results in Sect. \ref{sec:results} and 
%conclude in Sect. \ref{sec:conclusion}.

%%%%%%%%%%%%%%%%%%%%%%%%%%%%%%%%%%%%%%%%%%%%%

\section{Related Work}
\label{sec:related_work}
\textbf{Works on ACD.}
Most existing automatic ACD methods
model ACD as a classification task,
and their focuses are mostly on designing
useful features to represent clauses,
and selecting appropriate SL-based classifiers.
Widely used features include  structural, lexical, syntactic
and contextual features,
%(see, e.g. \cite{ig2014emnlp,moens2013argumentation})
and popular classifiers include SVM, naive Bayes,
decision tree and random forest.
For well-structured documents, using
these SL classifiers and conventional features 
leads to relatively good performances: for example,
SVM achieves .726 and .741 macro-$F_1$
in corpora consisting of persuasive essays
\cite{ig2014coling,ig2014emnlp} and legal documents
\cite{palau2009argumentation}, resp.
However, for some less well-structured texts, 
e.g. Wikipedia articles, % and product reviews, 
these methods
have significantly poorer performances:
% \cite{eckle2015role} investigated the 
% role of discourse markers
% for identifying argument components, and used 
% SVM to perform the classification.
\cite{levy2014claim_detection,lippi2015ijcai} 
report that in the task for detecting claims from Wikipedia
articles, only around .17 $F_1$ is achieved,
although they have tried different features 
(topic-dependent features and partial constituency 
trees, resp.) and different classifiers (logistic regression
and SVM, resp.).
%to perform this task.

% \cite{levy2014claim_detection} 
% used some topic-dependent features together 
% with the conventional features to represent
% clauses, and employed the logistic regression 
% algorithm to detect claims in Wikipedia
% articles; as an extension, \cite{lippi2015ijcai}
% removed the context-dependent features
% and only used partial constituency trees
% as features, and used a SVM to perform the detection; 
% however, both these techniques
% only achieved 17-18\% $F_1$.
% 
% and SL-based AD methods are rarely used in these
% texts \cite{wyner2012semi}. 
% 

Some works are devoted to using unsupervised-learning 
techniques to extract features.
\cite{lawrence2014lda} assume that clauses belonging to the 
same argument are likely to share the 
same topic; thus, they employ the
LDA-based topic modelling technique \cite{blei2003latent}
to extract each clause's topics and use
these topics as features.
\cite{nguyen2015extracting} use LDA to
extract the \emph{argument words}
(i.e. words used as argument indicators, e.g.
`think', `reason')
and \emph{domain words}
(i.e. terminologies commonly used within a certain
topic, e.g. `education', 'art'), and add 
indicator features for these words.
%Empirical results in these two works suggest that 
%the LDA-extracted features
%can improve the performance.
However, these features have only been tested
on small corpora constructed from well-structured documents
(the former is tested 
on documents obtained from a 19th 
century philosophical book, while the later
is tested on the persuasive essay corpus
proposed in \cite{ig2014coling}), 
and the computational expense of
LDA is high.
%for obtaining these features are high.

\textbf{Works on contextual information in ACD.}
HAs have been implicitly used in some SL-based ACD tools.
In \cite{ig2015emnlp}, ACD is modelled 
as a \emph{sequence tagging problem}, and 
SVM-HMM \cite{altun2003hidden} is used to solve 
this problem; SVM-HMM implicitly
considers type-C HAs during the labelling process.
Their technique achieves macro-$F_1$ between .185 and .304
in debate portal documents.
%However, they do not compare the performance 
%of SVM-HMM with any other classifiers, thus it
%is difficult to evaluate to what extent the HAs
%improve the performance.
Type-C HAs are arguably the de facto 
features used by RL-based NLP tools. 
In RL-based text summarisation techniques
\cite{ryang2012emnlp,rioux2014emnlp},
annotations of all preceding sentences,
made in the current round of scanning,
are included in the feature vector;
in RL-based dialogue generation
systems, e.g. \cite{pietquin2011sample,williams2007partially},
the full history of \emph{dialogue acts} in 
the current dialogue is used in the state 
representation. 
However, all these works do not
compare the performances of HAs-augmented
and HAs-free versions of their techniques,
thus fail to investigate to what extent the usage of 
HAs can improve performance.

As for other forms of contextual information,
in \cite{levy2014claim_detection}, the \emph{topic} of 
a document is used to build features to identify
claims. To be more specific, given a clause,
the similarity between this clause and the topic sentence
is used to decide whether this clause is 
a claim or not.
However, the importance of the topic information
is questionable, as \cite{lippi2015ijcai} report that similar performances
can be obtained without using the topic information.

%the topic information is only available
%in documents obtained from well-organised repositories
%(e.g. Wikipedia), and is unavailable for texts
%like hotel reviews.

%Since annotating arguments is seldom uncontroversial
%even for human experts, 
%Argument analysis is seldom uncontroversial, and thus
%annotating arguments in texts is 
%a challenging task even for human experts.
%Hence, the \emph{reliability of annotations} is
%an important metric to evaluate the quality of
%a corpus.

%%%%%%%%%%%%%%%%%%%%%%%%%%%%%%%%%%%%%

\section{Formulating ACD as a Sequential Decision Making Problem}
\label{sec:mdp_model}
\emph{Markov decision processes} (MDPs) 
are widely used mathematical models
for formulating sequential decision making problems.
%(c.f. \cite{littman1996algorithms}).
In this work,
we consider ACD formally as 
\emph{episodic MDPs}.
An episodic MDP is a tuple $(S,A,P,R,T)$. 
$S$ is the set of \emph{states}; a state is a representation
of the current status of the problem at hand. 
$A$ is the set of \emph{actions}. By performing an action $a$
in state $s$, the agent is transited to some new state $s'$
and receives a numerical reward $R(s,a)$,
where $R: S \times A \rightarrow \mathbb{R}$
is the \emph{reward function}.
$P(s'|s,a) \in [0,1]$ is the \emph{transition function}:
it gives the probability of moving from state $s$
to $s'$ by performing action $a$.
$T \subseteq S$ is the set of \emph{terminal states}:
when the agent is transited to a state $s \in T$, the current 
episode ends.
The components of our MDP-based ACD 
formulation are as follows:

\textbf{State set $S$.}
Each state $s$ represents a clause to be annotated.
Thus, we let $s$ be a feature vector, which includes 
not only the current clause's linguistic features, 
but may also include type-L and type-C HAs. 
We let $N_{a}$ denote the length of the conventional 
linguistic features,
$N_{l}$ and $N_{c}$ denote 
the window sizes for type-L and type-C
HAs, resp., and
$N$ denote the length of the state vector;
thus, $N = N_a + N_l + N_c$.
\iffalse
Note that, in our experiments with SL-based methods
(Sect. XX.XX),
we also use this feature vector to represent 
each clause, so as to guarantee the fairness
of our comparison.

By tuning $N_{a}$ ($N_{b}$, resp.), we can 
adjust the amount of type-L (type-C, resp.) HAs 
to be considered during the labelling process.
For example, by letting $N_{a} = 3$,
we consider the last round's annotation for 
the current clause and its neighbour clauses.
An illustration of the window size
is presented in Fig. XX. Thus, feature vectors
used by existing SL-based methods have $N_a=N_b = 0$.
\fi

\textbf{Action set $A$.}
Given a state $s$, performing action $a$ on 
$s$ means labelling the corresponding clause 
of $s$ as type $a$. Thus, each action in $A$ corresponds 
to a type of label. 

\textbf{Transition probability $P$.}
Given the above formulations of $S$ and $A$, 
$P(s'|s,a)$ gives the next clause $s'$ to be annotated
after the current clause's labelling finishes.
In other words, $P$ decides the sequence of labelling.
As such, we can use a short-hand notation
$P(s) = s'$ to indicate that, after $s$ is annotated,
no matter what its annotation is, the next 
clause to be annotated is $s'$;
in this work, for simplicity, we let $s'$ be the clause ensuing
$s$; investigating the effectiveness of other sequences of labelling
is left as a future work.

\textbf{Reward function $R$.}
$R(s,a)$ evaluates the goodness of annotating 
$s$ as type $a$. Thus, we let $R(s,a)$ be positive (negative, resp.)
if $a$ is (not, resp.) the same to the gold-standard annotation of
$s$. Note that function $R$ is known during the training phase 
%as the annotator knows the true annotations, 
but is unknown in the test phase. 
%Traditional planning techniques, e.g. dynamic programming,
%only manage to solve MDPs whose components are all known,
%and thus cannot be used to solve this MDP.

\textbf{Terminal states set $T$.}
We view labelling a document for one round as an 
episode. Thus, $s \in T$ if and only if 
$s$ corresponds to the last clause in a document. 

A policy $\pi: S \rightarrow A$ specifies
the action to take in each state.
RL amounts to algorithms
for obtaining the (near-)optimal policies for MDPs,
even some components (e.g. function $R$) of 
the MDP are unknown. 
To obtain the optimal policy, RL
maintains a \emph{Q-function}, which provides a 
quantitative evaluation of the current policy
being used.
Specifically,
given a policy $\pi$, its \emph{Q-function} 
$\qp(s,a)$ gives the
discounted sum of rewards that will be received by performing
action $a$ in state $s$ and following policy $\pi$ thereafter:
\begin{align}
\label{eq:q_def}
&\qp(s,a) = %\nonumber \\
%& 
E_{\pi}[r_0 + \gamma r_1 + \gamma^2 r_2 + \cdots],
\end{align}
where $r_t$ is the immediate reward received in time step $t$,
$E_{\pi}$ is the expectation operator with respect to policy $\pi$,
and $\gamma \in [0,1]$ is a real-valued parameter known as the 
\emph{discount factor}. 
%The goal of planning in a MDP is to find an 
%\emph{optimal policy} $\pi^*$,
%which maximises the discounted sum of future rewards;
%formally, $\pi^*$ satisfies:
%\begin{align}
%\label{eq:optimal_pi}
%\pi^*(s) = \arg \max_a \qps(s,a).
%\end{align}
%
%Since $\qps(s,a)$ considers all rewards received subsequent
%to the execution of $a$ in $s$, the optimal policy obtained using 
%Eq. \eqref{eq:optimal_pi} naturally takes the sequence of 
%states and actions into account; 
%this property of MDP greatly facilitates the representation
%of sequential information during the decision making,
%and this explains why MDP is among the most
%popular methods for modelling \emph{sequential decision
%making} problems \cite{Sutton98}.

\begin{figure}
\centering
\includegraphics[width=0.4\textwidth]{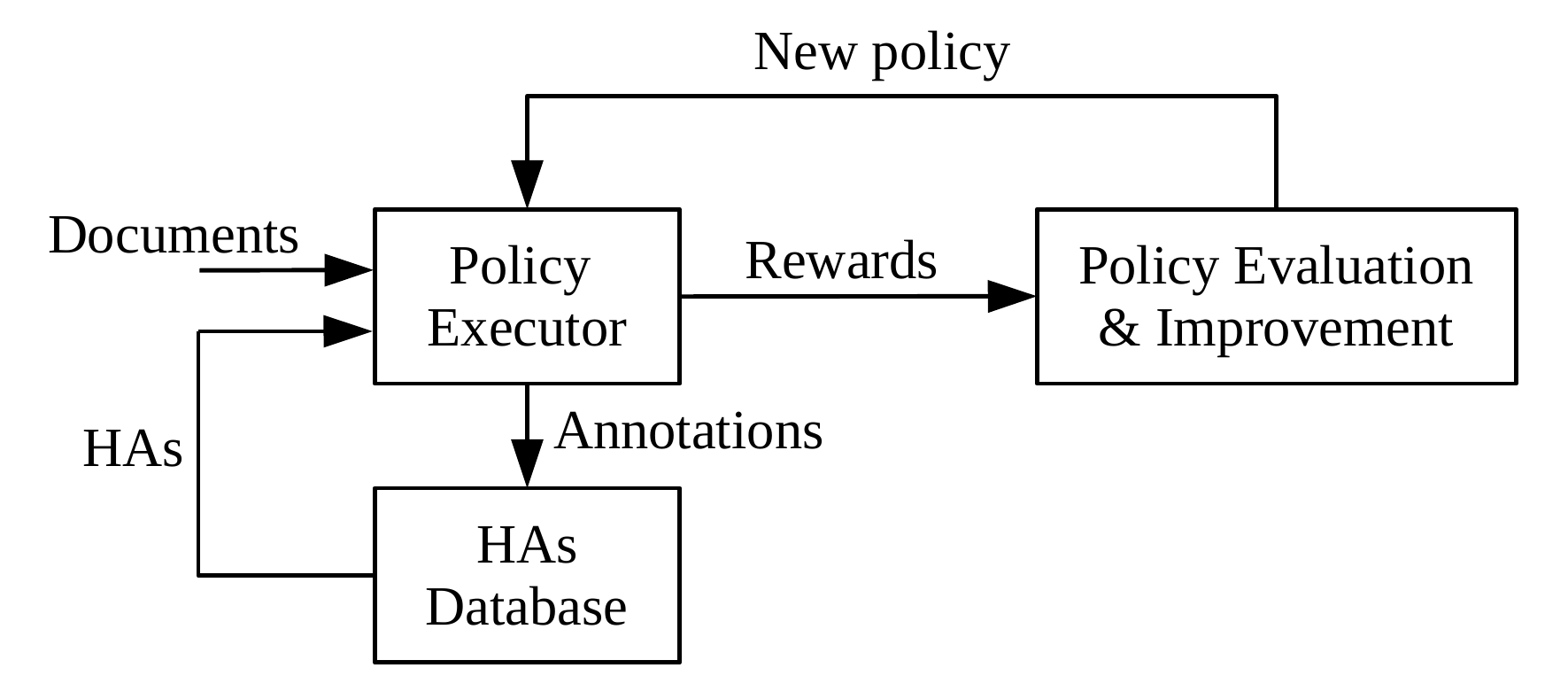}
\caption{The workflow of RL-based ACD. 
RL algorithm amounts to modules ``Policy Executor'' and 
``Policy Evaluation \& Improvement''.
Module ``HAs Database'' is used to store historical annotations (HAs).
\label{fig:rl_workflow}}
\end{figure}

In the ACD task, the RL agent proceeds as follows
to obtain the optimal policy:
the RL agent first uses some random policy to 
annotate the input documents for one round,
and collects the rewards (produced by reward function
$R$) during the labelling process; 
then the RL agent uses these rewards to 
build the Q-function of the policy, so as to
evaluate the goodness of the current policy and
derive an improved policy. The newly-obtained
policy is used to label the input documents for another
round, and the improvement cycle repeats until
the policy converges
(i.e. two consecutive policies are the same).
Most RL algorithms ensure that the converged policy is optimal.
Fig. \ref{fig:rl_workflow} illustrates the workflow
of RL-based ACD.

%Next, we will select a suitable RL algorithm
%for the MDP built in this section, and 
%present a RL-based AD framework.

%%%%%%%%%%%%%%%%%%%%%%%%%%%%%%%%%%%%%%%%%%%%%%%%%%5

\section{RL-based ACD Framework}
\label{sec:select_rl}
To select suitable RL algorithms for a MDP is not a 
trivial task, as RL algorithms fall into 
many different categories, each suitable for
certain types of MDPs.
We consider the following
two factors when we select RL algorithms for 
our MDP-based ACD formulation presented in Sect. 
\ref{sec:mdp_model}:
\begin{itemize}
\item 
\textbf{Data efficiency.}
Since there exist few
high-quality and large-scale ACD corpora
(c.f. \cite{lippi2016survey}),
we need to select RL algorithms with 
strong \emph{generalisation capabilities},
so as to learn the optimal policies with 
limited amount of training data.

\item 
\textbf{Computational efficiency.}
Obtaining the optimal policy usually requires 
many rounds of policy improvement. %\cite{Sutton98}.
Thus, the computational expense for each round
of improvement should be small enough. 
%otherwise the 
%overall computational burden will be overwhelming.
\end{itemize}

To strike a trade-off between the above 
two factors, we decide to use the 
\emph{least square policy iteration} (LSPI)
\cite{lagoudakis2003lspi} algorithm to 
solve our MDP-based ACD. LSPI is a 
\emph{model-based} RL algorithm, which 
%learn function $R$ during the learning process and 
can efficiently use the training data.
The computational complexity of LSPI increases
linearly with the growth of the sample size,
and some works have been proposed
to further reduce its complexity 
(e.g. \cite{geramifard2006ilstd,sutton2009fast}).

\begin{algorithm}[t]
\caption{The RL-based 
ACD method (one fold in the cross-validation).}
\begin{algorithmic}[1]
\label{alg:main}
\STATE{ \textbf{GIVEN:} } 
\STATE{the training set $D_{train}$,
test set $D_{test}$, discount factor $\gamma$, episode round $K$,
maximum scan round $J$}
\STATE{ \textbf{TRAIN:} }
\FOR{document $d$ in $D_{train}$}
  \FOR{each episode ($K$ episodes in total)}
    \STATE{
    invoke LSPI in each learning step, so as to evaluate
    and improve the current policy
%     obtain an annotation for each clause in $d$ by using some 
%     policy (e.g. $\epsilon$-greedy), obtain the $(s,a,r,s')$ tuple
%     and invoke \textbf{LSPI} learn from the tuple
    }
  \ENDFOR
\ENDFOR
\STATE{output the learnt policy $\pi^*$}
\STATE{ \textbf{TEST:} }
\FOR{document $d$ in $D_{test}$}
  \STATE{initialise $LastRoundAnnList$ as an empty list}
  \FOR{each round of scan ($J$ rounds at most)}
    \STATE{produce $AnnList$ for clauses in $d$, 
    using $\pi^*$ and considering $LastRoundAnnList$}
    \STATE{\textbf{break} if $LastRoundAnnList == AnnList$}
    \STATE{$LastRoundAnnList \leftarrow AnnList$}
  \ENDFOR
    \STATE{output and evaluate $AnnList$}
\ENDFOR
\end{algorithmic}
\end{algorithm}

The LSPI-based ACD framework is presented in 
Alg. \ref{alg:main}.
In the training phase (lines 4-8), 
the RL agent labels each document for 
$K$ rounds so as to obtain
the optimal policy; each round of labelling
is called an \emph{episode} (line 5).
The obtained policy is output after the training phase
finishes (line 9).

In the test phase (lines 11-19), 
since type-L HAs are used in the state representation
(see Sect. \ref{sec:mdp_model}),
the feature vector for the same clause can 
be different in different rounds (see
Fig. \ref{fig:has_types});
as a result, the algorithm needs to 
label the same document for multiple rounds
until the annotations converge,
i.e. the annotations in
the current round (stored in 
list $AnnList$, line 14) are the same to those
made in the last round (stored in $LastRoundAnnList$).
Once the annotations converge,
the algorithm breaks the loop (line 15)
and begins to label the next test document;
else, if annotations fail to converge 
in $J$ (an positive integer provided
a priori; see line 2) 
rounds of labelling, the algorithm outputs the 
annotations obtained in the final round (line 18).

Now we discuss the computational complexity 
of LSPI-based ACD.
Suppose there are $M$ clauses in the training set, and 
each document is labelled for $K$ rounds; also
remind that the state vector size is $N$ 
(see Sect. \ref{sec:mdp_model}). 
As such, the complexity of each episode (line 6)
using LSPI is $O(N^2)$, and the complexity
for obtaining the final policy (line 9) is $O(N^3)$
\cite{lagoudakis2003lspi};
thus, the overall complexity in the training
phase is $O(KMN^2 + N^3)$.

As for the complexity of SVM-based ACD,
again, we suppose that
there are $M$ clauses in the training set and 
each document is labelled for $K$ rounds.
For each clause, as its annotation can be
different in different rounds of labelling
(see Fig. \ref{fig:has_types}), 
each clause has $K$ different vector representations.
Thus, there are in total $KM$ input vectors
in the training phase.
%\footnote{The full workflow
%of SL-based AD using HAs-augmented features 
%is given in the supplementary material.}.
In line with most existing SL-based ACD 
(see Sect. \ref{sec:related_work}),
we select SVM with RBF kernel as the 
classifier, and
its complexity in the training phase
is between $O(N(KM)^2)$ and $O(N(KM)^3)$
(using LIBSVM 
\cite{chang2011libsvm}).
As for SVM-HMM \cite{altun2003hidden}, its 
complexity is no cheaper than standard SVM.
To summarise, the training complexity of
SVM/SVM-HMM is 
(at least) quadratic with the 
number of samples, while the complexity of
LSPI is linear with the number of samples; 
thus, LSPI-based ACD scales better
when applied to large-scale corpora,
and is more suitable for applications
with short feature vectors.

In the test phase, the complexity for 
computing the annotation for one clause 
%for one time 
is $O(N \cdot |A|)$ when using LSPI,  but 
is approximately $O(N^2)$ \cite{claesen2014fast}
when using SVM and SVM-HMM (with RBF kernel). 
Since the number of annotation types $|A|$
is usually much smaller than the vector
length $N$, the computational complexity
of LSPI is usually lower in the test phase.

%%%%%%%%%%%%%%%%%%%%%%%%%%%%%%%%%%%%%%%%%%%%%

\section{Datasets}
\label{sec:data}

When selecting corpora for testing our methods, 
we primarily consider the labelling quality
of the corpora, because
the corpora's quality heavily influences
the quality of the ACD tools trained on them
\cite{ig2015emnlp}.
%Most existing corpora asking multiple annotators
%independently label each document,
The \emph{inter-rater agreement} (IRA) score
is a widely used metric to evaluate the reliability
of annotations and quality of corpora. 
Fleiss' kappa \cite{fleiss1971kappa} is among the 
most widely used IRA metrics, because it
can compute the agreement between 
two or more raters, and
it considers the possibility of 
the agreement occurring by chance, thus giving
more ``robust'' measure than simple percentage agreement.
If the Fleiss' kappa score equals 1, it suggests 
the raters have ``perfect agreement''; the 
lower the score, the poorer the agreement.
In this work, %unless stated otherwise, 
all IRA scores reported are Fleiss' kappa values.

Since there exist few well-annotated and publicly available
argumentation corpora, we create our own
argumentation corpus.\footnote{Details of
the creation of our hotel corpus is presented 
in a separate paper, which is currently under review.} 
We randomly sampled
200 hotel reviews of appropriate length
(50 - 200 words) in the hotel review dataset
provided by \cite{wachsmuth2014review}.
We presented these hotel reviews on a crowdsourcing
platform, and asked five workers to independently 
annotate each review. 
Similar to \cite{wachsmuth2015sentiment}, we viewed
each sub-sentence as 
a clause. We asked the workers to label each clause 
as one of the following six categories:  
\begin{itemize}
\setlength\itemsep{0em}
\item \textit{major claim}: summarises the main opinion of a review;
\item \textit{claim}: an opinion on a certain aspect 
of a hotel;
\item \textit{premise}: a reason/evidence supporting a claim;
\item \textit{background}: an objective description 
that does not give direct opinions
but provides some background information;
for example ``this is my second 
staying at this hotel'', ``we arrived at 
at midnight''; 
\item \textit{recommendation}: a positive or 
negative recommendation for the hotel,
e.g. ``do not come to this place if you 
want a luxury hotel'', 
`I would definitely come to this hotel 
again the next time I visit London'; and
\item \textit{others}, for all the other clauses.
\end{itemize} 
A detailed annotation guide and some examples were
presented to the workers before they started their labelling.
We asked the workers to give one and only one major claim
for each hotel review, and informed them that a claim can have
no premises, but each premise must support some claim.
The annotating process lasts for 4 weeks, with 216 workers
in total participated.
We removed the annotations with obvious mistakes, 
and finally obtained annotations for 105 hotel reviews.
In total, the corpus contains 1575 sub-sentences and 14756 tokens;
some statistics are given in Table \ref{table:hotel}.
Since the IRA for type \textit{others}
is lower than 0.5, we manually checked and calibrated all
\textit{others} annotations.
Except for type \textit{others}, all types have IRA scores 
above 0.6, suggesting that the agreement is substantial
\cite{landis1977kappa}.

\begin{table}
\centering
\begin{tabular}{ | c | c c c c c c |}
 \hline
 & MC & Cl & Pr & Bg & Re & Others \\
 \hline
 Num & 105 & 345 & 180 & 75 & 71 & 120 
 \\ 
 IRA & 0.89 & 0.71 & 0.80 & 0.62 & 
 0.68 & 0.35 \\
 \hline

\end{tabular}
\caption{The number and IRA scores
for each type in the hotel reviews corpus.
\label{table:hotel}}
\end{table}

Another corpus we used to test our approach is the 
persuasive essays corpus proposed
in \cite{ig2016parsing}. 
This corpus contains 402 essays 
on a variety of different topics, and
it has three argument component types:
major claim, claim and premise;
the IRA scores for these three argumentative types 
are 0.88, 0.64 and 0.83, resp.;
however, the IRA for type \textit{others}
is not reported.

To the best of our knowledge, 
these two corpora are among the most well-annotated
argumentation corpora (in terms of IRA scores).
Some larger corpora, e.g.
the one in \cite{levy2014claim_detection},
have much lower IRA scores (.39);
the legal texts corpus
proposed in \cite{palau2009argumentation}
is not publicly available, and the  
web texts corpus proposed in \cite{habernal2014argumentation}
has relatively low IRA (below .50) for most 
argument component types.

\iffalse
The persuasive essays corpus contains 90 essays 
on a variety of different topics.
It adopts an argument model 
that has three types of argument components:
major claim, claim and premise. A major claim gives a 
short summary of a document;  
each essay has one and only one major claim. 
The class distribution and the
IAA scores are given in Table \ref{table:essays}.

\begin{table}
\centering
\begin{tabular}{ | c | c c c c |}
\hline
 & MC & Cl & Pr & N \\
 \hline
 Number & 90  & 429 
 & 1033 & 327\\
 IAA & 83.65 & 66.55 &
 71.31 & NA 
 \\
 \hline 
\end{tabular}
\caption{The number and IAA scores
for each type in the persuasive essays corpus.
`MC', `Cl', `Pr' and `N' stand for major claim, claim, premise and
non-argumentative clauses, resp. IAA for `N'
is not reported in their work.
\label{table:essays}}
\end{table}
\fi

%%%%%%%%%%%%%%%%%%%%%%%%%%%%%%%%%%%%%%%%%%%%%%%%%%%%%%%%%%5
\section{Experimental Settings and Results}
\label{sec:results}
%In this section, 
%we evaluate the performance of RL-based ACD
%and study the influences of HAs on different
%ACD techniques. 
In this section, 
we denote a HAs combination with type-L window size 
$i$ and type-C window size $j$ as a pair $(i,j)$.
Under each HAs combination setting,
%For preventing errors in model assessment due to a 
%particular data splitting (Krstajic et al., 2014), 
we used a repeated 10-fold cross-validation setup 
and ensured that clauses from the same document are not 
distributed over the train and test sets;
in addition,
we repeated the cross-validation 10 times,
which yields a total of 100 folds.
All results presented are average values over the 
100 folds. We 
%used the two-tailed t-test
%to compute the p-values, and 
let the significance level
be 0.05.
%In order to determine the macro F1 score, 
%we employ macro-averaging as proposed by Sokolova and Lapalme (2009, p. 430). 
As for the conventional linguistic features
(see Sect. \ref{sec:mdp_model}), 
we used exactly the same features
to those in \cite{ig2014emnlp}.
%in experiments on both corpora.
For model selection and hyper-parameter tuning,
we randomly sampled 25\% documents (from both
corpora) and performed 5-fold cross-validation.

\begin{table*}[t]
\centering
\begin{tabular}{| c | c c c c c c | c c c c c c c c |}
\hline
\multirow{2}{*}{ }
& \multicolumn{6}{c|}{In the essay corpus} & \multicolumn{8}{c|}{In the hotel 
corpus}
\\
\cline{2-15}
 & $F_1$ & Acc. & MC & Cl & Pr & Oth & $F_1$ & Acc. & MC & Cl & Pr & Re & Bg & Oth 
\\
 \hline 
 S & .650 & .719 & 
 .377 & \textbf{.462} & .792  & .965 & 
 .562 & .578 & .831 & .635$^\ddagger$ 
 & .413 & .633 & .654 & .085$^\ddagger$
 \\
 M & .683 & .705 & 
 .517$^\dagger$ & .442 & .770 & \textbf{.966} & 
 .582 & .556 & .902$^\dagger$ & .531 
 & .506$^\dagger$ & .652 & .661 & .020
 \\
 \hline
 R & .656 & .697 & .525$^\dagger$ & .413 & 
 \textbf{.852$^{\dagger \ddagger}$} & .833 & 
 .521 & .549 &  .752 & .623$^\ddagger$ & 
 .410 & .655 & .484 & .054 
 \\
 H & \textbf{.696$^\dagger$} & \textbf{.759$^{\dagger \ddagger}$}
 & \textbf{.539$^\dagger$}  & 
 .459 & .845$^{\dagger \ddagger}$ & .925 & 
 \textbf{.638$^{\dagger \ddagger}$} & \textbf{.647$^{\dagger \ddagger}$}
 & \textbf{.941$^{\dagger \ddagger}$} & \textbf{.704$^{\dagger \ddagger}$} & 
 \textbf{.509$^\dagger$} & \textbf{.719$^{\dagger \ddagger}$} & 
 \textbf{.673} & 
 \textbf{.204$^{\dagger \ddagger}$}
 \\
\hline
\end{tabular}
\caption{Averaged performances of some ACD methods.
% All results are in percentage.
%averaged over 20 repeated 10-fold (for essay)
%or 12-fold (for hotel) CVs. `$F_1$' stands for the macro-$F_1$ score, and
%result for each type is its $F_1$ score.
% `Best' and `Worst' stand for RL-based AD method with 
% the best and worst HAs combinations (in terms of macro-$F_1$),
% resp.
In the left-most column, S, M, R and H stand for SVM,
SVM-HMM, HAs-free RL and HAs-augmented RL, resp.
$F_1$ and Acc. stand for averaged macro-$F_1$ and accuracy, resp.
Results for each type are averaged $F_1$ scores.
$^\dagger$: significant improvement 
over SVM; $^\ddagger$: significant improvement 
over SVM-HMM.
%best result is in boldface, results with no significant
%gaps comparing to the best performance
%are in italic, and results significantly inferior than `RL'
%are underlined.
\label{table:sig_results}
}
\end{table*}

\subsection{Baselines} \label{subsec:baselines}
We select SVM and SVM-HMM as our baselines,
because these two algorithms are among the most
widely used and best-performing
algorithms to build ACD tools
(see Sect. \ref{sec:related_work}).
As for the algorithm implementations,
we used LIBSVM \cite{chang2011libsvm} for SVM and 
a revision of SVM$^{struct}$ \cite{joachims2009cutting} for
SVM-HMM. %we finally used to RBF kernels in both SVM and SVM-HMM,
%as it outperforms the linear and polynomial kernels. 

Although SVM-HMM considers type-C HAs, it ignores
type-L HAs and it does not consider HAs explicitly.
For these reasons, and also for ensuring the 
fairness of comparison between SL-
and RL-based ACD tools, we also test the performances
of SVM and SVM-HMM using the HAs-augmented features.
%as described in Sect. \ref{sec:mdp_model}.
%Note that when using HAs-augmented features,
%each document will be classified for $K$ times,
%because the same clause may have different feature
%vectors in different rounds (see Fig. \ref{fig:has_types}
%and the complexity discussion in Sect. \ref{sec:mdp_model}).
We try all HAs combinations
from (0,0) to (9,5) in both baseline algorithms,
and find that using HAs-augmented
features does not result in significant changes 
on the performances of SVM and SVM-HMM
in both corpora. 
We have tried using HAs-augmented features in 
some other SL algorithms (J48 decision tree, 
naive Bayes and random forest provided in
WEKA \cite{hall2009weka}), and we make 
similar observations.
These results suggest that, most existing SL-based ACD
tools can hardly take advantage of HAs to improve
their performances, either through the 
implicit way (e.g. SVM-HMM, which implicitly
considers type-C HAs) or the explicit way
(i.e. directly augmenting HAs into the feature vector).

As for the relative goodness of SVM and SVM-HMM,
these two baseline approaches have comparable
performances in both corpora:
consider the best performances achieved by
SVM and SVM-HMM (presented in  
the first two rows in Table \ref{table:sig_results});
in both corpora,
although macro-$F_1$ of SVM-HMM is marginally
higher than those of SVM,
the accuracy of SVM is higher than that of SVM-HMM,
and the gaps between their accuracy
and macro-$F_1$ scores are all insignificant.
%Since SVM-HMM implicitly uses type-C HAs,
%this observation suggests that type-C HAs 
%can indeed improve the performance of SL-based ACD,
%but how to effectively use type-L HAs in 
%SL-based ACD remains an open question that 
%we leave for future work.
%Detailed results of SVM and SVM-HMM 
%(under the best-performing HAs combinations) 
%are presented in .
%and the performance distributions 
%are presented in Fig. \ref{fig:distributions}.
%T-tests suggest that, for each algorithm, 
%using HAs-augmented features 
%does not lead to significant performance changes;
%however, the performance (with respect to macro-$F_1$) 
%of SVM-HMM is superior than that of SVM. 

\iffalse
\begin{figure}
\centering
\subfigure[Hotel corpus]{\label{subfig:svm-hotel}
\includegraphics[width=0.22\textwidth]
{pics/hotel-oneHot-boxplot.pdf} }
\subfigure[Essay corpus]{\label{subfig:svm-essay}
\includegraphics[width=0.18\textwidth]
{pics/essay-oneHot-boxplot.pdf} }
\caption{Macro-$F_1$ distributions
for SVM and SVM-HMM
using different HAs combinations.
% Results presented here are macro $F_1$ values
% of a single 10-fold (or essay) or 12-fold
% (for hotel) CV.
\label{fig:distributions}}
\end{figure}
\fi

\subsection{Results}
\label{subsec:results}
%For both HAs-free and HAs-augmented RL,
%we let $K=10$, $J=10$ and $\gamma=0.4$ 
%(see line 2 in Alg. \ref{alg:main} for 
%these parameters).
%and 
%the initial policy we used is the \emph{random policy},
%which randomly selects actions in each state. 

\begin{figure}[t]
\centering
\subfigure[Hotel corpus.]{\label{subfig:rl-hotel}
\includegraphics[width=0.24\textwidth]
{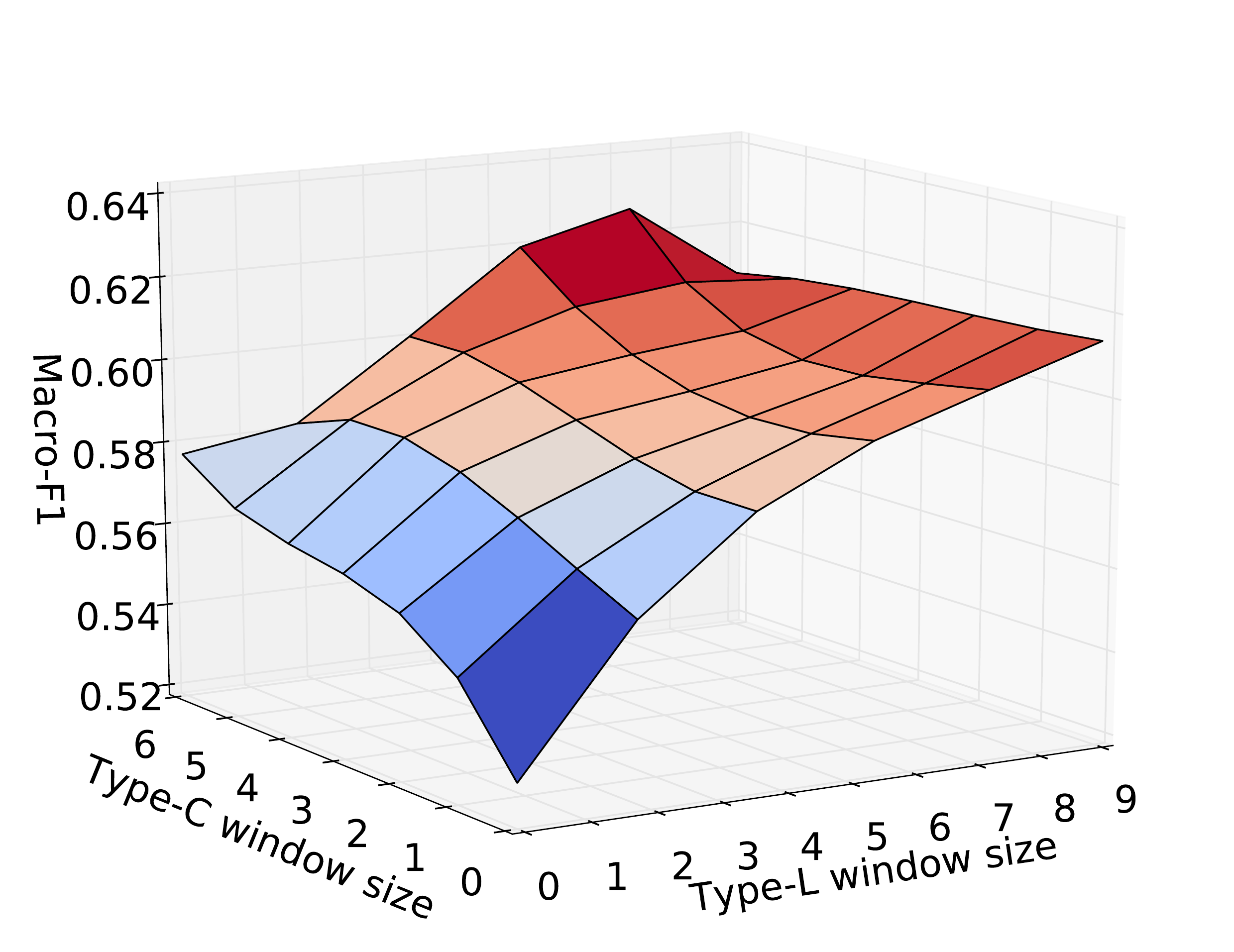} }
\hspace{-10pt}
\subfigure[Essay corpus.]{\label{subfig:rl-essay}
\includegraphics[width=0.23\textwidth]
{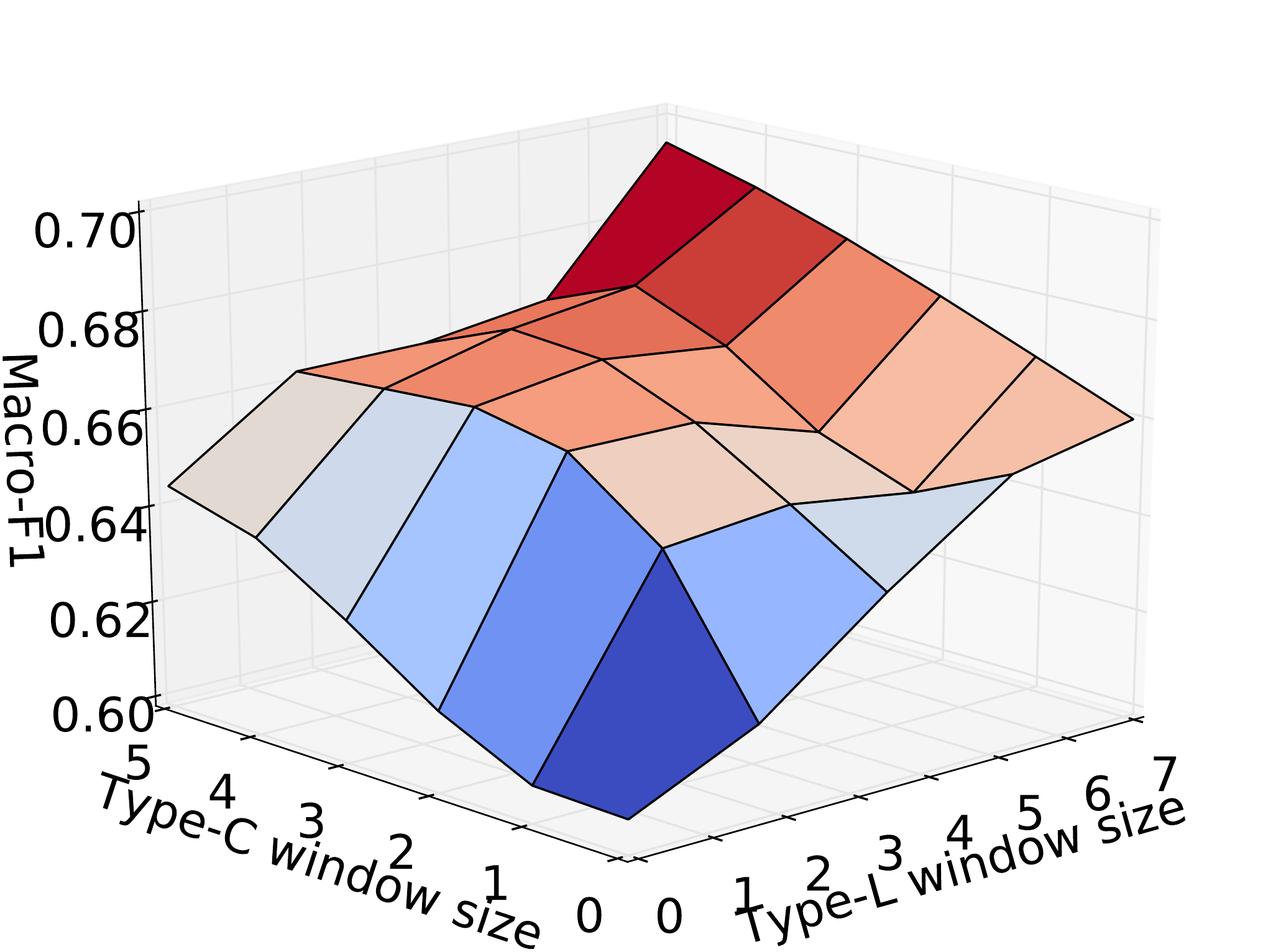} }
\caption{Influences of HAs for RL-based ACD.
% Results presented here are macro $F_1$ values
% of a single 10-fold (or essay) or 12-fold
% (for hotel) CV.
\label{fig:rl-wind-size}}
\end{figure}

First, we study the influences of HAs
on RL-based ACD.
The performances of RL-based ACD using different HAs
combinations are presented in Fig. \ref{fig:rl-wind-size}.
We can see that, in both corpora,
the worst performances are obtained
when no HAs are used, and the performances
increase almost linearly with 
the growth of the type-L and type-C window sizes.
To evaluate the significance of the improvement,
we performed t-tests between performances 
at $(0,0)$ (no HAs are used), $(7,0)$ 
(type-L is used to the maximum and no type-C is
used), $(0,5)$ (type-C is used to the maximum
and no type-L is used) and $(7,5)$ 
(both types of HAs are fully used);
results suggest that the performance
at $(0,0)$ is significantly inferior than 
the performances obtained in the other three
settings, and the performance at $(7,5)$
is significantly superior than all the
other performances.
These observations indicate
that, both type-L and type-C can significantly
improve RL's performance in the ACD task,
and the two types of HAs can be used together.
%and the best performances are achieved at
%$(7,5)$ in both corpora.  
As for the relative importance
of type-L and type-C, since the performances' 
growth rate along the type-L and type-C dimensions 
are almost the same, we believe that the relative
importance of these two types of HAs are comparable
and their influences on the performances are 
independent. 

Second, we compare the performances of RL-based
ACD and the baseline approaches. Main results
are presented in Table \ref{table:sig_results}. 
The results for HAs-augmented RL (the last row in
the table) are obtained using the HAs combination
$(7,5)$.
We make two key observations from these results:
\begin{itemize}
  \item
    HAs-free RL underperforms the baseline 
    approaches in both corpora, but the performance gaps
    are mostly marginal and insignificant.
    To be more specific, except for the 
    gap between the macro-$F_1$ score of HAs-free RL (.521) and 
    SVM-HMM (.582) in the hotel corpus, 
    all other gaps between HAs-free RL's
    performance (namely accuracy and macro-$F_1$)
    and those of the baseline approaches are not 
    statistically significant.
   \item 
    HAs-augmented RL outperforms the baseline approaches 
    in both corpora, and the performance gaps are mostly 
    substantial and significant.
    Specifically, except for the gap between 
    the macro-$F_1$ score of HAs-augmented RL (.696) and
    SVM-HMM (.683) in the essay corpus, 
    all other gaps between HAs-augmented RL's
    performance (accuracy and macro-$F_1$) and those of the baseline approaches are
    significant.
\end{itemize}
The reason that HAs-free RL underperforms
SVM/SVM-HMM is because the RL algorithm we use
(namely the LSPI algorithm; 
see Sect. \ref{sec:select_rl} and Alg. \ref{alg:main})
has much weaker ``expressiveness''  than SVM/SVM-HMM:
LSPI employs a linear function to evaluate Q-function
(see Sect. \ref{sec:mdp_model} and Eq. \eqref{eq:q_def})
thus, when the Q-function is complex, LSPI can hardly
provide a precise estimation of the Q-function;
since Q-function is used to derive annotation policies, 
the poor estimation of Q-function 
harms RL's performance.
In contrast, SVM and SVM-HMM use RBF kernels
to perform the classification, which has much
stronger expressiveness than LSPI's linear function.
Thus, we believe that by using more sophisticated 
RL algorithms, e.g. the \emph{kernel-based RL} algorithms
\cite{taylor2009kernelized,ormoneit2002kernel}
%\emph{RL with Gaussian processes}
%\cite{engel2005gprl,deisenroth2015gaussian},
and the recently proposed \emph{deep RL}
\cite{mnih2015human},
the performance of RL-based ACD can be substantially
improved (at the price of higher computational complexity though).

\subsection{Discussion and Error Analysis}
\label{subsec:error_analyses}

To obtain further insights into how the
usage of HAs
improves the performance of RL, we look into
the confusion matrices of each algorithm
and manually investigate some misclassified
cases. In both corpora, we find the biggest
error source is the misclassification between
premises and claims: for example,
in the essay corpus, 607 out of 1506 claims
are mis-classified as premises by SVM-HMM;
in the hotel corpus, 88 out of 180 premises
are mis-classified as claims by HAs-free RL.
Similar observations are also reported 
in \cite{ig2014emnlp}, and we believe that
ignoring contextual information (including HAs)
is a major factor leading to these misclassifications,
as illustrated in Example 1.% in Sect. \ref{sec:intro}.
We find that this problem is considerably 
mitigated by HAs-augmented RL: 
from Table \ref{table:sig_results} we can see
that, in the essay corpus, HAs-augmented RL's
performance for type \textit{premises} 
leads baselines' performances by around 8\%;
in the hotel corpus, HAs-augmented RL's
performance for type \textit{claim} 
outperforms baselines' performances by over 10\%.
As a concrete example, HAs-augmented RL
correctly labels \textcircled{2} to \textcircled{5}
in Example 1 as premises, while the other approaches
fail.

In the hotel corpus, another major error source
is the mis-classification of type \textit{others}:
from the right-most column in Table \ref{table:sig_results}
we can see that the $F_1$ score for type
\textit{others} is the lowest among all types'
performances. We believe the reason is that,
it is even challenging for human annotators to identify
clauses of type \textit{others}; this can be seen from the fact
that the IRA score for \textit{others}
is the lowest (0.35; see Table \ref{table:hotel}).
As a concrete example, consider the 
following review excerpt in our hotel corpus:

\begin{quote}
\textbf{Example 2}:
\textcircled{1}
The rooms come in different sizes. 
\textcircled{2}
The two other families we were traveling with had larger rooms 
-- 
\textcircled{3}
see if you can book something larger (especially if traveling with kids). 
\end{quote}

%In the crowdsourcing-based annotating process,
In the crowdsourcing platform,
3 workers labelled \textcircled{3} as \textit{others}
while the other 2 workers labelled it as \textit{recommendations}.
We think the reason is that, when 
considering \textcircled{3} alone, it looks like a  
\textit{recommendation}; but when considering all the three
clauses, \textcircled{3} is more like \textit{others}. 
Thus, HAs are important for identifying 
type \textit{others}, and
this may explain why HAs-augmented RL
outperforms the other three ACD techniques
by such a big margin for type \textit{others}:
when we let both type-L and type-C windows larger than
3, HAs-augmented RL can successfully label 
\textcircled{3} as \textit{others}, while the other 
approaches label it as \textit{recommendation}.

\section{Conclusion}
\label{sec:conclusion}
In this work, we novelly propose a RL-based
ACD technique, and study the influences
of HAs therein.  Empirical results on two
corpora suggest that,
using HAs can significantly improve RL-based
ACD's performance, and the HAs-augmented
RL's performance is significantly superior than
those of the state-of-the-art SL-based ACD techniques.
To the best of 
our knowledge, this is the first work systemically
studying the influences of HAs and the applicability 
of RL in the ACD task.
Future work includes studying the influences 
of some other contextual information, e.g. 
linguistic features of surrounding clauses, 
in SL- and RL-based ACD methods, and 
studying the applicability of RL for some other
sub-tasks in argumentation mining, 
e.g. argument relation prediction.

%HAs is only  part of the contextual information (CI)
%people considered in their labelling process;
%other information includes linguistic features
%of some surrounding clauses or some long-distance
%but important clauses (e.g. the first or last clause).
%The role of other forms of CI in AD in different 
%ML-based AD methods is worth further investigations. 
%In additions, RL can be used in some other sub-tasks
%in AM, for example the argument relation prediction 
%task.

% In current AM pipeline (c.f. \cite{lippi2016survey}), 
% argument detection and 
% argument relation prediction (ARP)
% are formulated as two separate tasks; 
% however, human annotators simultaneously 
% consider both argument component types and 
% their relations when the perform the argumentation
% mining task. Since the annotation for each 
% clause may influence future annotations
% on relations, we believe that RL can be used 
% to perform the AD and ARP task simultaneously.
% 

\bibliographystyle{named}
\bibliography{general_short}

\end{document}